\documentclass{article}
\usepackage{ijcai18}

\usepackage{times}
\usepackage{xcolor}
\usepackage{soul}
\usepackage[utf8]{inputenc}
\usepackage[small]{caption}

\usepackage [autostyle, english = american]{csquotes}
\MakeOuterQuote{"}
\usepackage{url}
\usepackage{amsmath}
\usepackage{amsfonts}       
\usepackage{nicefrac}       
\usepackage{microtype}
\usepackage{graphicx}
\usepackage{subfigure}
\usepackage{color}
\usepackage{booktabs}

\usepackage{multirow}
\usepackage{multicol}
\usepackage{tabularx,ragged2e,booktabs,caption}

\newcommand{\train}{\text{train}}
\newcommand{\test}{\text{test}}
\newcommand{\lap}{\text{lap}}
\newcommand{\ed}{\text{ed}}

\title{Drug Similarity Integration Through Attentive Multi-view Graph Auto-Encoders}

\author{
Tengfei Ma$^{1,2}$, 
Cao Xiao$^{1,2}$, 
Jiayu Zhou$^3$,
Fei Wang$^4$
\\ 
$^1$ IBM Research, $^2$ MIT-IBM Watson AI Lab\\
$^3$ Computer Science and Engineering, Michigan State University \\
$^4$ Weill Cornell Medical School, Cornell University  \\
tengfei.ma1@ibm.com, cxiao@us.ibm.com, jiayuz@msu.edu, few2001@med.cornell.edu
}

\begin{document}

\maketitle

\begin{abstract}
Drug similarity has been studied to support downstream clinical tasks such as inferring novel properties of drugs (e.g. side effects, indications, interactions) from known properties. The growing availability of new types of drug features brings the opportunity of learning a more comprehensive and accurate drug similarity that represents the full spectrum of underlying drug relations. However, it is challenging to integrate these heterogeneous, noisy, nonlinear-related  information to learn accurate similarity measures especially when labels are scarce. Moreover, there is a trade-off between accuracy and interpretability. In this paper, we propose to learn accurate and interpretable similarity measures from multiple types of drug features. In particular, we model the integration using multi-view graph auto-encoders, and add attentive mechanism to determine the weights for each view with respect to corresponding tasks and features for better interpretability. Our model has flexible design for both semi-supervised and unsupervised settings. Experimental results demonstrated significant predictive accuracy improvement. Case studies also showed better model capacity (e.g. embed node features) and interpretability.
\end{abstract}

\section{Introduction}\label{sec:introduction}

The rapidly evolving technologies have made it easier to collect multiple types of drug data and thus opened new opportunities for computational drug discovery research and drug safety studies. The study of drug similarity paves the foundation for these research since similar structural, molecular and biological properties often relate to similar drug indications or adverse effects~\cite{jamia12}. In literature, drug similarity has been computed using molecular structure data~\cite{AAAI1714292}, interaction profile data~\cite{Fokoue:2016:PDI:2872518.2890532}, as well as side-effect information~\cite{zhao2014jamia,Liue28}. 

Recently, there has been a growing interest in learning improved drug similarity from multiple types of drug features. For example, ~\cite{li2015inductive} proposed an inductive matrix completion method to combine multiple data sources and help predict the unknown side effects. ~\cite{zhang2015label} proposed an integrative label propagation algorithm to infer clinical side effects from multiple sources with considering high-order similarity. Results from these pilot studies show that combined similarity measures are usually more informative and robust to noise.  These methods could be summarized into four major categories: the nearest neighbor method, the random walk based approaches, the unsupervised, and the multiple kernel learning methods. Section~\ref{sec:related work} provides more details of the related literature.

Despite potential benefits, when learning from multiple biomedical data sources, significant challenges arise from the simultaneous handling of the following issues: 1) different types of features have different levels of associations with targeting outcomes. For example, drugs' structural similarity could have more influence on their interaction profiles than drugs' indication similarity do; 2) the underlying relations of biomedical events (e.g., two drugs interact to cause a side effect) are often nonlinear and complex over all types of features~\cite{Imming06}; 3) data quality (e.g. lack of label, noise in the data) also creates challenges for similarity learning, and 4) a model that captures complex drug relations is often be very complex and lacking interpretability.

To address the aforementioned challenges, we consider each type of drug feature as a view and learn integrated drug similarity using multi-view graph autoencoders (GAE). In particular, we model each drug as a node in the drug association network and extend the graph convolutional networks (GraphCNN)~\cite{kipf2016semi} to embed multi-view node features and edges.
Across views, we use attentive view selection scheme to enable nonlinear multi-view fusion and make the learning more interpretable and adaptive to data. By such embedding, we learn drug similarity and use them to predict outcomes (e.g., drug-drug interactions). In addition, for the setting where we would like to integrate multiple drug similarity graph without knowing any features, we propose an alternative transductive learning method based on treating labels as latent variables. The proposed models not only improve prediction performance, but also have the following benefits.
\begin{itemize}
\itemsep0em 
\item \textit{Intepretable and adaptive multiview fusion}: To model the heterogeneous relevance among different views with targeting tasks, in our similarity integration, we use attentive model to fuse multiple views. The attentive view selection scheme generates task-wise feature relevance, by which we could learn interpretable similarity measures. Also the learned similarity would be more adaptive to the underlying data, thus is more accurate.

\item\textit{Transductive prediction using unlabeled data}: Labels are expensive to acquire, and often very scarce for new drugs. By developing an auto-encoder structure, whose reconstruction loss could be seen as a regularization term that explicitly models the information of graph structure, we efficiently leverage the unlabeled data for accurate predictions. 

\item \textit{Robust-to-noise}: The proposed methods inherit the advantage of autoencoders and can extract representations that are relatively stable and robust to the noise in the data, e.g. in the drug-drug interaction prediction case, sometimes unseen interactions might not indicate no interaction. The proposed methods effectively reduce the negative impacts caused by these ``positive unlabeled'' samples.
\end{itemize}

\section{Related Work}\label{sec:related work}

Our work addresses the problem of multiview similarity integration. To our best knowledge, current approaches mainly could be summarized as below. \\

\noindent \textbf{The nearest neighbor methods}  that make predictions based on majority cases among neighbors. To name a few, ~\cite{Zhang:2016:PPS:2868724.2868741}, ~\cite{Zhang2017}, and ~\cite{zhang2015label}. However, as pointed out by ~\cite{zhang2015label}, most of these existing methods only utilize first-order similarity to construct neighborhood and do not consider transitivity of similarities. 

\noindent  \textbf{The random walk methods} (e.g., label propagation in ~\cite{zhang2015label} and ~\cite{Wang:2010:LBD:1933307.1934495}) that leverage the assumption that data points occupying the same manifold are very likely to share the same semantic label, and then aim to propagate labeling information from labeled data points to unlabeled ones according to the intrinsic data manifold structures collectively revealed by a large number of data points. These methods can handle nonlinear relations and perform transductive learning with scarce labeled data. However, these models have fixed loss functions, hence lack of flexibility in modeling various problem settings. 

\noindent \textbf{The unsupervised methods} For example, in ~\cite{naturemethod14} and \cite{Angione2016}, the authors construct an integrative network to fuse multiple similarity networks via an iterative scaling approach. In~\cite{10.1371/journal.pone.0152792},the authors integrated feature ranking and feature variation as feature weights for weighted similarity fusion. These unsupervised methods have good flexibility, however without any supervision, unreliable results could be generated. 

\noindent\textbf{The multiple kernel learning (MKL) methods} such as ~\cite{pmlr-v15-zhuang11a}. MKL were further extended to integrate heterogeneous data in ~\cite{McFee:2011:LMS:1953048.1953063}, however, most existing methods are often limited to convex integration.

\section{Background}

Over the past few years, several graph-based convolutional network models emerged for inducing informative latent feature representations of nodes and links. For example,~\cite{kipf2016semi} proposed a new graph convolutional network (GraphCNN) that learns node embeddings based on node features and their connections, which could be used in node classification. Specifically, given an undirected graph with nodes $X$ and adjacency matrix $A$, a multi-layer neural network is constructed on the graph with the following layer-wise propagation rule:
$$
\boldsymbol{H}^{(l+1)} = f\left(\tilde{\boldsymbol{D}}^{-\frac{1}{2}}\tilde{\boldsymbol{A}}\tilde{\boldsymbol{D}}^{-\frac{1}{2}} \boldsymbol{H}^{(l)}\boldsymbol{W}^{(l)}\right)
$$
where $\tilde{\boldsymbol{A}} = \boldsymbol{A} + \boldsymbol{I}_N$ is the adjacency matrix with added self-connections,  $\boldsymbol{D}$ is a diagonal matrix such that $\boldsymbol{D}_{ii} = \sum_j \tilde{\boldsymbol{A}_{ij}}$, $\boldsymbol{W}^{(l)}$ is a layer-specific parameter matrix, $H^l$ is the node representation in the $l^{th}$ layer, and $f$ is an activation function (e.g. ReLU or sigmoid).
Later, ~\cite{2017arXiv170306103S} and ~\cite{kipf2016variational} extended GraphCNN and proposed a graph auto-encoder (GAE) using GraphCNN for both node classification and link prediction tasks. However, their model only reconstructs the edges, and cannot work on unseen data.  In the following, we will make further extension based on ~\cite{kipf2016variational} and ~\cite{2017arXiv170306103S} in terms of reconstructing both links and node embeddings and allowing for inductive prediction.

\section{Method}

In this paper, we consider each type of drug feature as a view. For view $u\in\{1,\cdots,T\}$, we construct a graph by modeling each drug as a node and the similarity between two nodes as an edge. We denote node feature embeddings as $\boldsymbol{Z}^{(u)} $ and use similarity matrix $\boldsymbol{A}^{(u)}$ to represent the pair-wise similarity between drugs on that view. Given $T$ different views, the task of multi-view similarity integration is to derive an integrated node embedding $\boldsymbol{Z}$ and similarity matrix $\boldsymbol{A}\in \mathbb{R}^{n*n}$ across all views. 

\subsection{Similarity Integration with Attentive Multi-view Graph Autoencoders}
\label{sec:method}

\begin{figure*}[!h]
\centering
\begin{tabular}{c c c}
\includegraphics[width=0.3\textwidth]{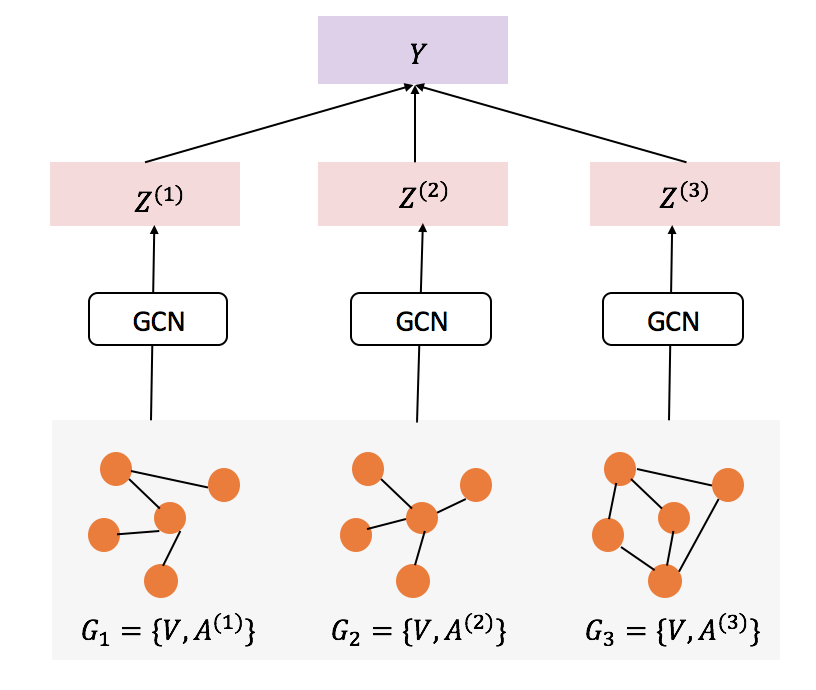} & 
\includegraphics[width=0.3\textwidth]{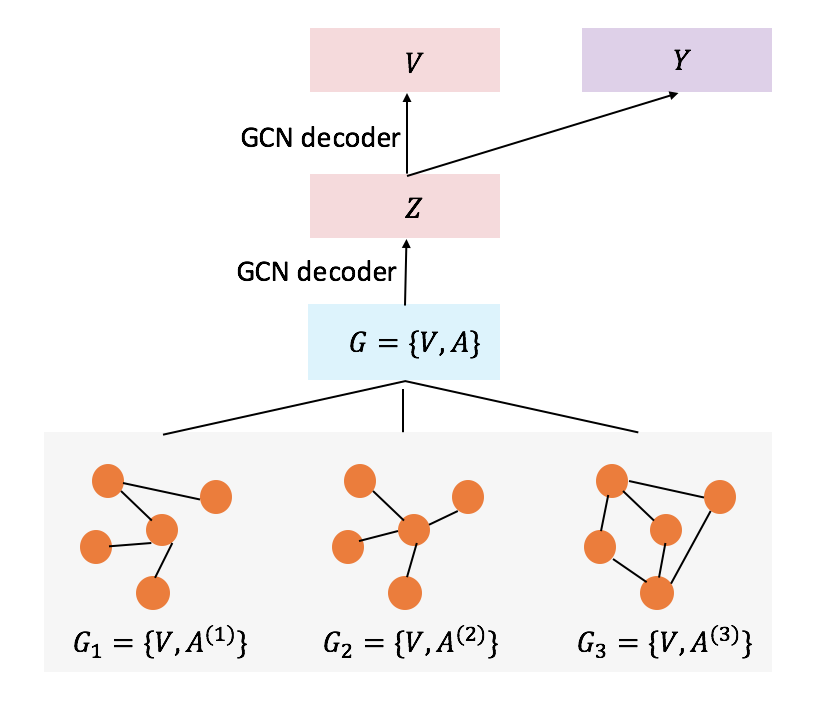} & 
\includegraphics[width=0.3\textwidth]{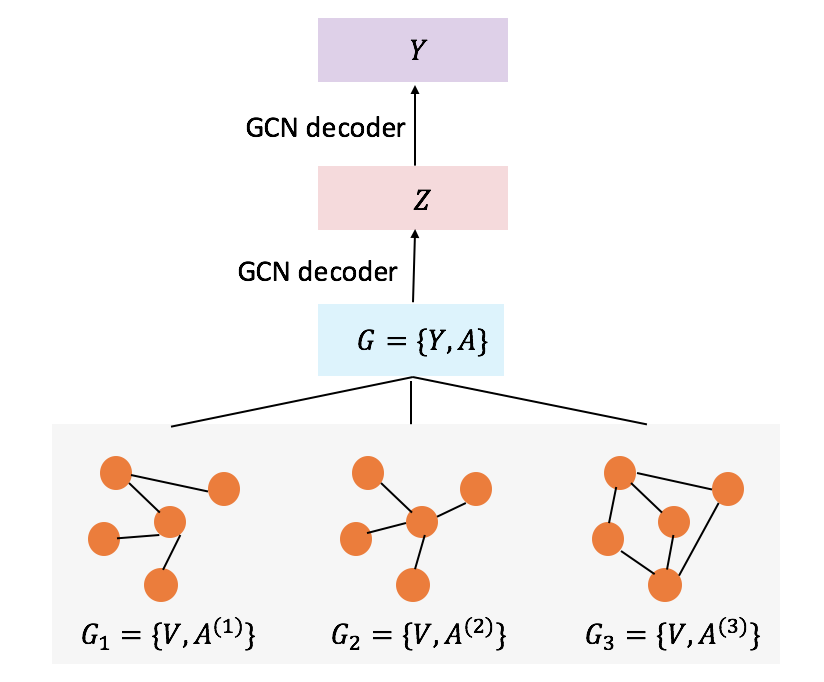} \\
(a) & (b) &(c)
\end{tabular}
\caption{The illustration of GraphCNN for link prediction: (a) The basic GraphCNN structure with multiple similarity matrices. (b) Semi-supervised graph auto-encoder based on GraphCNN. (c) Transductive graph auto-encoder.}
\label{fig:basecase}
\end{figure*}

\noindent\textbf{Basic GraphCNN Structure with Multiple Views} 
For each view $u$, we set $\tilde{\boldsymbol{A}}^{(u)} = \boldsymbol{A}^{(u)} + \boldsymbol{I}_N$ and diagonal matrix $\boldsymbol{D}^{(u)}$ where $\boldsymbol{D}^{(u)}_{ii} = \sum_j \tilde{\boldsymbol{A}}^{(u)}_{ij}$, then we use a two-layer GraphCNN to get the node embeddings $\boldsymbol{Z}^{(u)}$ using Eq.~(\ref{eq:embedding}).
\begin{eqnarray}
\label{eq:embedding}
\boldsymbol{Z}^{(u)} &=& f(\boldsymbol{X}^{(u)}, \boldsymbol{A}^{(u)}; \boldsymbol{W}^{(u)}_1) \\\nonumber
&=& \text{Softmax}\left( \boldsymbol{\hat{A}}^u \text{ReLU}(\boldsymbol{\hat{A}}^u \boldsymbol{X}^{(u)} \boldsymbol{W}^{(u)}_0)  \boldsymbol{W}^{(u)}_1 \right),
\end{eqnarray}
where $\boldsymbol{\hat{A}}^u = {\boldsymbol{D}^{(u)}}^{-1/2} \tilde{\boldsymbol{A}}^{(u)} {\boldsymbol{D}^{(u)}}^{-1/2}$,  $\boldsymbol{W}^{(u)}_0$ and $\boldsymbol{W}^{(u)}_1$ are weight matrices.
Given the node embeddings $\boldsymbol{Z}^{(u)} $ on each view $u$, we concatenate the embedding from each view $\boldsymbol{Z}^{(u)}$ to get a new representation of the node $\boldsymbol{Z}$. The prediction between two nodes $x_i$ and $x_j$ could be done by a sigmoid function $y(x_i, x_j) = \sigma(\boldsymbol{z}_i^T \boldsymbol{W} \boldsymbol{z}_j)$ with a matrix parameter $\boldsymbol{W}$. The structure of this method is shown in Figure~\ref{fig:basecase}(a).\\

\noindent\textbf{Similarity Matrix Fusion} Instead of concatenating node embeddings in different views, we can also first get an integrated similarity matrix and construct only one graph for all views. In this single graph, the nodes features are fixed for all views. And the similarity fusion could be simply done as follows: considering the complexity of normalization, to fuse similarity we first normalize all similarity matrices $\boldsymbol{A}^u$ to get $\boldsymbol{\hat{A}}^u$, and then aggregate all similarity matrices to get a comprehensive one as the adjacency matrix of the graph: $\boldsymbol{\hat{A}} = \frac{1}{T}\sum \alpha_u \boldsymbol{\hat{A}}^u$, where $\{\alpha_u\}$ are mixing weights for different similarity matrices. Following the structure in \cite{kipf2016semi}, we use a one-layer GraphCNN to encode the nodes in our graph: 
\begin{eqnarray}
\boldsymbol{Z} = f(\boldsymbol{X},\boldsymbol{\hat{A}}) =  \text{Softmax}(\boldsymbol{\hat{A}} X\boldsymbol{W}_0) 
\end{eqnarray}
After that, we decode the embedding back to the original feature space 
\begin{eqnarray}
\boldsymbol{X}^\prime = f^\prime(\boldsymbol{Z},\boldsymbol{\hat{A}}) = \text{Sigmoid} (\boldsymbol{\hat{A}} \boldsymbol{Z}\boldsymbol{W}_1)
\end{eqnarray}
If we do not have any labels for the nodes, the objective function is the loss of the auto-encoder in Eq.~\ref{eq:loss1}.
\begin{equation}
L_{\ed} = \sum ||\boldsymbol{X} - \boldsymbol{X}^\prime||^2
\label{eq:loss1}
\end{equation}
In this case, our framework could be regarded as an unsupervised multi-graph fusion and embedding method. 
The derived similarity matrix $\boldsymbol{\hat{A}}$ can be used for other tasks as well, such as node clustering. \\

\begin{figure}[!h]
\centering
\includegraphics[width=0.5\textwidth]{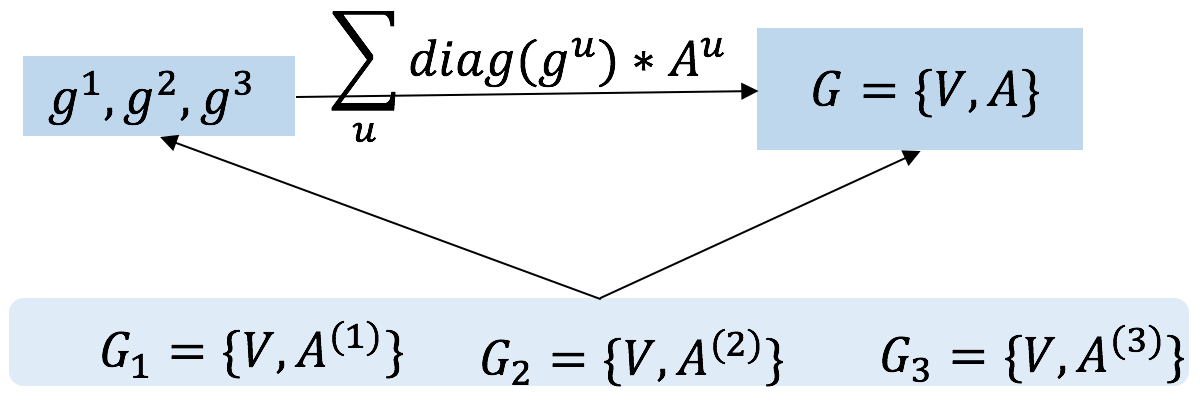} 
\caption{The illustration of attention view selection scheme.}
\label{fig:gatedview}
\end{figure}

\noindent\textbf{Attentive View Selection}  In practice, the fusion of each view could be nonlinear, while the weights of features in each view need to be decided by both the data and the targeting tasks.
To allow for such a flexibility, in this section we extend the mixing scheme by 
fusing features from different views with attention mechanism, where weights of features are determined by corresponding inputs. The attentive view selection scheme is illustrated in Fig.~\ref{fig:gatedview}.

Assume we have adjacency matrix $\boldsymbol{A}^u$ for view $u$, we assign attention weights $\boldsymbol{g}^u \in \mathbb{R}^{N*N}$ to the graph edges, such that the integrated adjacency matrix becomes $\sum_u \boldsymbol{g}^u \odot \boldsymbol{A}^u$, where $\odot$ is the element wise multiplication. For each view, we first project the original adjacency matrix to an unnormalized matrix $\boldsymbol{g}^{\prime u} = \boldsymbol{W}^u \boldsymbol{A}^u +\boldsymbol{b}^u$, and then normalize them over different views to get the attention weights $\boldsymbol{g}^u$. In practice, the graph is often large, thus there will be too many parameters for the attention calculation if we use a fully connected $N*N$ attention matrix. To reduce the complexity, we alternatively employ a diagonal attention matrix. To be specific, we limit $\boldsymbol{g}^u$ to be a vector, and form the weighted similarity matrix by $diag(\boldsymbol{g}^u) * \boldsymbol{A}^u$. In this way, the size of parameters (i.e. $\boldsymbol{W}^u$ and $\boldsymbol{b}^u$ ) is reduced from $N*N$ to $N$. And the attentive similarity matrix is generated as follows:
$\boldsymbol{g}^{\prime u} = \boldsymbol{W}^u \boldsymbol{A}^u +\boldsymbol{b}^u$,
where $\boldsymbol{W}^u, \boldsymbol{b}^u \in \mathbb{R}^{1*N}$.

Then we normalize them to get the attention weights for each position $i$:
$\boldsymbol{g_i} = (g^1_i,...,g^T_i) = \text{Softmax}(g_i^{\prime 1}, ..., g_i^{\prime T})$, and $\boldsymbol{g^u} = (g^u_1,...g^u_N)$ is then used to induce the final similarity matrix $\boldsymbol{A} = \sum_u{\text{diag}(\boldsymbol{g}^u) * A^u}$, where $*$ is the matrix multiplication, $\text{diag}(\boldsymbol{g}^u)$ is a diagonal matrix of $g^u$ as its diagonal value. After we get the new attention based similarity matrix $\boldsymbol{A}$, we can use the same framework as \ref{sec:semi_gae} and \ref{sec:trans_gae}.

\subsection{A Semi-supervised Extension Given Partial Labeled Data}\label{sec:semi_gae}

The graph auto-encoder (GAE) structure could be further extended to a semi-supervised setting when we have labels for some of the nodes in the graph (in our case drugs). We could keep the auto-encoder framework unchanged, and predict the labels on training data using a network $h$: $h(\boldsymbol{Z}) = \text{Softmax}(\boldsymbol{W}_h \boldsymbol{Z} + \boldsymbol{b}_h)$. The prediction loss of $h$ is formulated by Eq.~(\ref{eq:loss2}).
\begin{eqnarray}
L_{\train} = L(\boldsymbol{Y_{\train}}, h(\boldsymbol{Z}_{\train})) = \sum_{y\in \boldsymbol{Y}_{\train}} \sum_{y^\prime \in h(\boldsymbol{Z}_{\train})} -y \ln{y^\prime}
\label{eq:loss2}
\end{eqnarray}
This new model then integrates the two loss functions as its objective function 
\begin{equation}
L = L_{\train}+ \lambda L_{\ed}
\label{eq:loss3}
\end{equation}
Compared to a generic neural network, which generally contains only the $L_{train}$, Eq. ~(\ref{eq:loss3}) can be seen as adding an auto-encoder loss as the regularization term $L_{ed}$. In a graph based semi-supervised learning framework, the graph Laplacian regularization is often used as the regularization to capture the graph structure. 
$L = L_{\train} + \lambda L_{\lap}$, where 
$L_{lap} = \sum_{i,j} \boldsymbol{A}_{i,j}||h(x_i)-h(x_j)||^2$.
Our objective function replaces the second term with the reconstruction loss of the GAE, which also explicitly models the graph structure information.

\subsection{Transductive Learning using Test Labels as Variables}\label{sec:trans_gae}
Sometimes when we only have the graph structure of the similarity matrix but no node features, although we could model them using one-hot representation as in \cite{kipf2016semi} (for details, see Appendix A.1 in \cite{kipf2016semi}), such embedding is typically not efficient. More importantly decoding the embedding vectors to the one-hot vectors cannot gain much information. This motivates us to develop another scheme to extend the previous introduced graph auto-encoder to improve learning given no node feature.

Instead of using the original node features or one-hot node vectors in GAE, we consider an alternative way: we use the training labels (i.e. DDI links of each node) as inputs and reconstruct them using the same GAE structure as in Figure~\ref{fig:basecase}(c). So the graph auto-encoder would output the predicted links $\boldsymbol{Y}^\prime = f^\prime(f(\boldsymbol{Y}, \boldsymbol{\hat{A}}), \boldsymbol{\hat{A}})$.

Moreover, if we consider similarities as graph edges, the labels of the test nodes would also impact the decoding of the training nodes. So we employ a transductive method to use the test labels as additional latent variables. The predicted labels are formulated as follows:
\begin{eqnarray}
\{\boldsymbol{Y}^\prime_{\train}, \boldsymbol{Y}^\prime_{\test}\} = f^\prime(f(\{\boldsymbol{Y}_{\train}, \boldsymbol{Y}_{\test}\}, \boldsymbol{\hat{A}}), \boldsymbol{\hat{A}})
\label{eq:trans_model}
\end{eqnarray}
i.e. $\boldsymbol{Y}^\prime_{\test}$ is a function of $\boldsymbol{Y}_{\test}$ when $\boldsymbol{Y}^\prime_{\train}$, $\boldsymbol{Y}_{\train}$ and $\boldsymbol{\hat{A}}$ are known. The objective function of this model is then given by:
\begin{eqnarray*}
\min_{Y_{test}}{L} = ||\boldsymbol{Y}^\prime_{\train} - \boldsymbol{Y}_{\train}||^2 + ||\boldsymbol{Y}^\prime_{\test}-\boldsymbol{Y}_{\test}||^2+\mu ||\boldsymbol{Y}_{\test}||^2,
\label{eq:trans_obj}
\end{eqnarray*} 
where $\mu ||\boldsymbol{Y}_{\test}||^2$ is a regularization term which enforces stability of the solutions. Thus after inference from the training data, we can get the optimal neural network parameters as well as the latent variables $Y_{test}$.

\section{Experiment}

Detecting adverse drug-drug interaction (DDI), a modification of the effect of a drug when administered with another drug, is one of the clinically important applications as DDIs result in large amounts of fatalities per year and incur huge morbidity and mortality related cost of $\sim\$177$ billion annually \cite{Giacomini2007}. Making use of multiple drug characterizations in similarity computation is critical since drugs could have heterogeneous similarity in different feature dimensions, e.g. drugs that have similar chemical structures could have very different therapeutic target and thus result in different DDI mechanism.

\subsection{Data Sources}

\noindent\textbf{Binary Prediction of the Occurrence of DDIs}: For the first data set, we will integrate multiple similarity graph (without node feature) to predict whether there will be interaction between a new pair of drugs. 
In the data, we have the following views: 1) DDI: The known labels of DDIs are extracted from the Twosides database~\cite{tatonetti2012data}, including 645 drugs and 1318 DDI events, in total $63473$ distinct pairs of drugs associated with DDI reports. 2) Label Side Effect: Drugs' side effects extracted from SIDER database ~\cite{SIDER4} are considered one type of features, including $996$ drugs and $4192$ side effects. We call this view as “Label Side Effect” by the convention in ~\cite{zhang2015label}. 3) Off-Label Side Effect: Drugs confounder-controlled side effects from OFFSIDES dataset are considered another type of features, including $1,332$ drugs and $10,093$ side effects. 4) Chemical Structure: Drug structure features (i.e. chemical fingerprints) are structural descriptors of drugs. In our study, we generate drug structure features with the extended-connectivity fingerprints with diameter 6 (ECFP6) using the R package “rcdk \cite{rcdk}”. The features are hashed binary vectors of 1,024-bit length, of which each bit encodes the presence or absence of a substructure in a drug molecule. We used Jaccard index to compute similarities between all the fingerprints.\\


\noindent\textbf{Multilabel Prediction of Specific DDI Types}: For the second data, we integrate multiple type of drug views to predict specific interaction types among $1301$ candidate types for new drug pairs. In the data, we have $222$ drugs and the following views: 1) Drug Indication: The drug indication data of dimension $1702$ is downloaded from SIDER \cite{SIDER4}. It is originally generated from MedDRA database, which is a widely used clinically-validated international medical terminology. 2) Drug chemical protein interactome (CPI): The CPI data  from ~\cite{repo} provides an important measure about how much power a drug needs to bind with its protein target. Its dimension is $611$. The similarity of CPI is calculated using the RBF kernel. 3) Protein and nucleic acid targets (TTD): For each drug, we associate its multiple protein and nucleic acid targets information and generate features of dimension $207$. These entries are extracted from the Therapeutic Target Database (TTD) ~\cite{ttd}. 4) Chemical Structure: The chemical structure features are extracted in the same way as in dataset 1, except that we chose "pubchem" fingerprint instead, whose feature dimension is $582$.


\subsection{Implementation and Evaluation Strategy}

\noindent\textbf{Proposed Model}: We implement the proposed model with Tensorflow 1.0 \cite{tensorflow2015-whitepaper} and trained using Adam with learning rate 0.01 and early stopping with window size 30. We optimized the hyperparameter for SemiGAE on validation data and then fixed for all GAE models: 0.5 (dropout rate), 5e-4 (L2 regularization) and 64 (\# of hidden units). For GCN models, we have the second layer and the number of the hidden units in the second layer is set as $32$. \\

\noindent\textbf{Baseline}: In addition, we implemented the following four baselines for comparison: 
\begin{itemize}
\item Nearest Neighbor (NN): We implemented the NN method in ~\cite{vilar2012drug}. It identifies novel DDIs by using the nearest neighbor similarity to drugs involved in established DDIs.
\item Label Propagation (LP): We considered the LP model in ~\cite{zhang2015label} as a baseline. The LP method propagates the existing DDI information in the network to predict new DDIs, and could also integrate multiple similarity matrices in the network.
\item GraphCNN:  For single view, we use the same structure as the nonprobabilistic GAE model in~\cite{kipf2016variational}. We consider the DDI links as edges and form the adjacency matrix. For multiple views, we linearly integrate all similarity matrices as well as the training DDI links.
\item Multiple Kernel Learning (MKL): For MKL, we used the python "Mklaren" library \cite{DBLP:journals/corr/StrazarC16}. We set $\sigma=30$ for RBF and degree $p=3$ for polynomial kernel. We only applied MKL on Data 2 since Data 1 does not have features for all views.
\end{itemize}
For all models, we use Tanimoto coefficient (TC) to calculate similarity except for CPI. For CPI, we measure drug similarity using RBF kernel. For all methods (except NN which already has the similar procedure in its method), following the procedures in ~\cite{Zhang:2016:PPS:2868724.2868741}, after getting the predicted labels using our model, we calculate the probability of that drug $i$ interacts with drug $j$ by $score_{ji}=score_{ij}=Y_{ij}+Y_{ji}$.\\

\noindent\textbf{Evaluation}: In evaluation, we adopted strategies in~\cite{zhang2015label} and randomly selected a fixed percentage (i.e., $25\%$ and $50\%$) of drugs, and moved all DDIs associated with these drugs for testing. For the data not in testing, we train on $90\%$ and perform validation and model selection on $10\%$ of the drugs. For testing data, we repeated the hold-out validation experiment $50$ times with different random divisions of the data, and reported the mean and the standard deviation of the area under the receiver operating characteristic curve (ROC-AUC) as well as the area under the precision-recall curve (PR-AUC) over the 50 repetitions. In the ROC and PR analytics, we utilized DDI interactions from TWOSIDES as reference positives, and the complement set as reference negatives.

\subsection{Results}

Table ~\ref{tab:data1} and ~\ref{tab:data2} compare the performance of the proposed models against baselines on both datasets. From the tables we can see for both single view and multi-view, the proposed models significantly outperform baselines. Also, the multiview models generally outperform corresponding single view models since our integrations provide more comprehensive measures of drug similarity. With adding attention mechanism, the relevant types of features receive more weights in similarity integration. 

In addition, we observed that the attentive semi-supervised GAE (AttSemiGAE, the model of Section~\ref{sec:semi_gae}) often achieves the best ROC-AUC, which is due to the embedding of node features. This advantage is more obvious on Dataset 2 than Dataset 1, since for Dataset 2, we have node features on all views, while most views in Dataset 1 have no node feature. For Dataset 1, due to the lack of node feature in most views, the attentive transductive GAE (attTransGAE, the model of Section~\ref{sec:trans_gae}) gains better PR-AUC thanks to transductive learning from tests labels and adaptive weight learning.

\vspace{-0.1in}

\begin{table*}[ht]
\centering
  \caption{Predicting Binary DDI Outcomes on Dataset 1.}
  \label{tab:data1}
  \begin{tabular}{l|l|cc|cc}
  \hline
  \multicolumn{6}{c}{Using Single View }\\\hline
 \multirow{2}{*}{} & \multirow{2}{*}{\small Methods}  & \multicolumn{2}{c}{Test Split ($25\%$)}& \multicolumn{2}{c}{Test Split ($50\%$)} \\
    & & ROC-AUC & PR-AUC& ROC-AUC & PR-AUC\\\hline
\multirow{3}{*}{\small Baselines} & \small  NN & \small $0.693 \pm 0.025$  & \small $0.504 \pm 0.043$ & \small $0.681 \pm 0.027$ & \small $0.498 \pm 0.042$  \\
& \small  LP & \small $0.769 \pm 0.013$ & \small \boldmath{$0.648 \pm 0.028$} & \small $0.752\pm0.018$& \small $0.632 \pm 0.035$ \\
 & \small GraphCNN & \small $0.723 \pm 0.028$  & \small $0.536 \pm 0.050$ & \small $0.717 \pm 0.039$& \small $0.528 \pm 0.076$\\\hline\hline
\multirow{2}{*}{\small Proposed}& \small  SemiGAE  & \small \boldmath{$0.788 \pm 0.014$}  &\small $0.640 \pm 0.036$  & \small \boldmath{$0.785 \pm 0.012$}& \small \boldmath{$0.637 \pm 0.035$} \\
 & \small  TransGAE & \small $0.786 \pm 0.013$ & \small $0.644 \pm 0.033$& \small $0.776 \pm 0.013$& \small $0.630 \pm 0.039$ \\\hline
 \multicolumn{6}{c}{Using Multiple Views}\\\hline
 \multirow{2}{*}{\small Baselines} & \small  LP & \small $0.768 \pm 0.013$& \small $0.644 \pm 0.028$ & \small $0.751 \pm 0.018$& \small $0.629 \pm 0.036$ \\
& \small  GraphCNN & \small $0.696 \pm 0.061$ & \small $0.515 \pm 0.066$ & \small $0.636 \pm 0.047$ & \small $0.457 \pm 0.064$\\\hline\hline
\multirow{2}{*}{\small \bf Proposed}&  \small AttSemiGAE & \small \boldmath{$0.798 \pm 0.013$} & \small $0.655 \pm 0.032$ & \small \boldmath{$0.791 \pm 0.017$}& \small $0.642 \pm 0.037$\\
  & \small AttTransGAE & \small $0.785 \pm 0.012$ & \small \boldmath{$0.687 \pm 0.023$} & \small $0.780 \pm 0.016$& \small \boldmath{$0.678 \pm 0.031$}\\
  \hline
  \end{tabular}
\end{table*}


\begin{table*}[ht]
\centering
  \caption{Predicting Specific DDI Types (Multiple Outcomes) on Dataset 2.}
  \label{tab:data2}
  \begin{tabular}{l|l|cc|cc}
  \hline
  \multicolumn{6}{c}{Using Single View }\\\hline
 \multirow{2}{*}{} & \multirow{2}{*}{\small Methods}  & \multicolumn{2}{c}{Test Split ($25\%$)}& \multicolumn{2}{c}{Test Split ($50\%$)} \\
    & & ROC-AUC & PR-AUC& ROC-AUC & PR-AUC\\\hline
\multirow{3}{*}{\small Baselines} & \small  NN & \small $ 0.627\pm 0.043 $  & \small $0.594 \pm 0.078$ & \small $ 0.594\pm 0.033 $ & \small $0.554 \pm 0.061$  \\
& \small  LP & \small $0.773 \pm 0.025$ & \small \boldmath{$0.670 \pm 0.052$} & \small $0.747\pm 0.028$& \small \boldmath{$ 0.650\pm 0.053$} \\
 & \small GraphCNN & \small $0.738 \pm 0.047 $  & \small $0.594 \pm 0.080 $ & \small $0.698 \pm 0.090 $& \small $ 0.583 \pm 0.102 $\\
 \hline\hline
\multirow{2}{*}{\small Proposed}& \small  SemiGAE  & \small \boldmath{$0.798 \pm 0.029$}  &\small $0.661 \pm 0.059$  & \small \boldmath{$0.784 \pm 0.028 $}& \small $ 0.649 \pm 0.059 $ \\
 & \small  TransGAE & \small $0.790 \pm 0.028$ & \small $ 0.661\pm 0.068$& \small $0.770 \pm 0.031 $& \small $ 0.633\pm 0.080$ \\\hline
  \multicolumn{6}{c}{Using Multiple Views}\\\hline
  \multirow{2}{*}{\small Baselines} & \small  LP & \small $0.774 \pm 0.025$& \small $0.672 \pm 0.052$ & \small $0.748 \pm 0.028$& \small $0.653 \pm 0.055 $ \\
& \small  GraphCNN & \small $0.601 \pm 0.067$ & \small $0.526 \pm 0.120$ & \small $0.578 \pm 0.067$ & \small $0.526 \pm 0.108 $\\
& \small MKL & \small $0.766 \pm 0.030 $  & \small $0.650 \pm 0.061 $ & \small $0.724 \pm 0.026 $& \small $ 0.586  \pm 0.066 $\\
\hline\hline
\multirow{2}{*}{\small \bf Proposed}&  \small AttSemiGAE & \small \boldmath{$0.802 \pm 0.029$} & \small \boldmath{$0.678 \pm 0.060 $} & \small \boldmath{$0.786 \pm 0.030$}& \small \boldmath{$0.662 \pm 0.064$}\\
  & \small AttTransGAE & \small $0.782 \pm 0.026$ & \small $0.670 \pm 0.058$ & \small $0.764 \pm 0.025$& \small $0.652 \pm 0.061 $\\
  \hline
  \end{tabular}
\end{table*}


\subsection{Case Studies}

\textbf{Understanding the Major Source of Similarity} When two drugs cause similar DDIs, such a similarity could be induced by various mechanisms. For example, drugs that prolong the QT interval, drugs that are CYP3A4 inhibitors, or drugs that alter another drug's metabolism via cytochrome P450 interactions or changes in protein binding, etc ~\cite{jyp}. Better understanding the major DDI mechanism would benefit us from developing actionable insights to identify proper ways to prevent DDIs. In this paper, adding attention mechanism enhances the interpretability of the models and could potentially provide understanding of the underlying DDI mechanism.

\begin{table}[ht]
\centering
\caption{Attention Weights for Selected DDIs}
\label{tab:interpret}
\begin{tabular}{l|c|c|c|c|c}
\hline
DDI Type  & AUC & chem.  & indi. & TTDS & CPI\\
\hline
\small Chest Pain &  \small $0.772$ & \small $0.151$ & \small $0.303$ & \small $0.144$ & \small $0.402$\\
\small Insomnia & \small $0.755$ & \small $0.380$ & \small $0.261$ & \small $0.078$ & \small $0.291$\\
\small Aching Muscles & \small $0.774$ & \small $0.117$ & \small $0.301$ & \small  $0.283$ & \small $0.299$\\
\hline
\end{tabular}
\end{table}

Table.~\ref{tab:interpret} reports several selected DDIs and the weights of each views as predicted using AttSemiGAE. For example, the DDI "chest pain" has good prediction AUC, and the views "CPI" and "indication" both have more impact on the predictions than other views. We consult domain expert, and find it in line with domain knowledge. Many DDI cases of chest pain are due to particular drug overdose, such as Venlafaxine and Mirtazapine ~\cite{Nachimuthu12}, which could be prescribed together to treat depression. However, the co-use of them could cause overdose thus prolong the QT interval via chemical protein interactome (CPI), and eventually cause chest pain. For another DDI "insomnia", one major mechanism is the interaction between cytochrome P450 (CYP) inducers (e.g. Rifampicin) and Hypnosedatives. Insomnia happens when the cytochrome P450 (CYP) inducers significantly induce the metabolism of the newer hypnosedatives and decreased their sedative effects \cite{CNS}. Such a process was caused by the bindings of chemical structures with proteins. In the results, the weights for "pubchem" and "CPI (compound-protein binding)" are much higher than the rest, in line with knowledge.\\

\noindent\textbf{Importance of Multiview Feature Integration}: We also examined how feature integrations across multiples views of features could help provide more accurate measures of drug similarity.

For example,  Acyclovir (Pubchem ID 2022) and Ganciclovir (Pubchem ID 3454), having medium level similarity in  indication and TTDS, since Acyclovir is used for treating herpes simplex virus infections and shingles but Ganciclovir is mainly used in more severe Cytomegalovirus diseases and AIDS. However, they both are analogues of 2'-deoxyguanosine and have very high structural similarity (0.961 measured using "Pubchem" fingerprint). The high structural similarity lead to many common DDIs shared by the two drugs according to the groundtruth. Our proposed model adaptively gives more weight to the structural similarity and computes an integrated similarity score at $0.682$, but label propagation (LP) fails to capture such heterogeneous influences and yields an integrated score at only $0.551$, which is an underestimate comparing with the groundtruth.

Similar examples include the similarity between Alprazolam (Pubchem ID 2118) and Estazolam (Pubchem ID 3261) as well as the similarity between Alprazolam (Pubchem ID 2118) and Triazolam (Pubchem ID 5556). The two pairs of drugs have quite low indication similarity, however, they all interact when in combined use with CYP3A4 inhibitors such as Cimetidine, Erythromycin, Norfluoxetine, Fluvoxamine, Itraconazole, Ketoconazole, Nefazodone, Propoxyphene, and Ritonavir. The combined uses will delay the hepatic clearance of Alprazolam, Estazolam or Triazolam, which then cause accumulation and increased severity of side effects from these drugs. The proposed model account for the feature heterogeneity and weight more on the chemical structural feature and CPI feature, leading to integrated similarity at $0.682\sim 0.720$, while other methods considered each views homogeneously and the resulting similarity is often low at $0.551 \sim0.630$.

\vspace{-0.1in}

\section{Conclusion}

In this paper, we proposed a set of Graph Auto-Encoder based models that perform multi-view drug similarity integration with attention model to perform view selection. The nonlinear and adaptive integration not only offers superior predictive performance but also interpretable results. We extended these GAE models to semi-supervised/transductive settings and predict the unknown DDIs. Experimental results on two real-world drug datasets demonstrated the performance and efficacy of our methods. Future works include expansion along the line of data or model. Data-wise, we could try on a larger drug database to fully exploit of the power of deep learning without overfitting. Model-wise, we will pursuit directly computing integrated similarity across multiple views without the need for calculation similarity for each view first.

\section*{ Acknowledgments}
The work of Fei Wang is supported by NSF IIS-1750326 and IIS-1716432. The work of Jiayu Zhou is funded by NSF IIS-1749940, IIS-1615597 and by ONR under N00014-17-1-2265.

\newpage

\bibliographystyle{named}
\bibliography{ijcai18}

\end{document}